\documentclass[twoside,english,american]{elsarticle}
\usepackage[T1]{fontenc}
\usepackage[latin9]{inputenc}
\usepackage{mathrsfs}
\usepackage{slashed}
\usepackage{amstext}
\usepackage{amsthm}
\usepackage{stackrel}
\usepackage{graphicx}

\makeatletter
\theoremstyle{plain}
\newtheorem{thm}{\protect\theoremname}
\theoremstyle{definition}
\newtheorem{defn}[thm]{\protect\definitionname}
\theoremstyle{definition}
\newtheorem{example}[thm]{\protect\examplename}

\journal{Journal}

\usepackage{amsfonts}
\usepackage{amsmath}
\usepackage{amssymb}

\makeatother

\usepackage{babel}
\addto\captionsamerican{\renewcommand{\definitionname}{Definition}}
\addto\captionsamerican{\renewcommand{\examplename}{Example}}
\addto\captionsamerican{\renewcommand{\theoremname}{Theorem}}
\addto\captionsenglish{\renewcommand{\definitionname}{Definition}}
\addto\captionsenglish{\renewcommand{\examplename}{Example}}
\addto\captionsenglish{\renewcommand{\theoremname}{Theorem}}
\providecommand{\definitionname}{Definition}
\providecommand{\examplename}{Example}
\providecommand{\theoremname}{Theorem}

\begin{document}

\begin{frontmatter}{}

\title{Multiple-criteria Heuristic Rating Estimation}

\author[rvt]{Anna K\k{e}dzior}

\ead{kedzior@wms.mat.agh.edu.pl}

\author[focal]{Konrad Ku\l akowski\corref{cor1}}

\ead{kkulak@agh.edu.pl}

\cortext[cor1]{Corresponding author}

\address[rvt]{AGH University of Science and Technology, Faculty of Applied Mathematics,
Krak?, Poland}

\address[focal]{AGH University of Science and Technology, Faculty of Electrical Engineering,
Automatics, Computer Science and Biomedical Engineering, Krak?, Poland
}
\begin{abstract}

One of the most widespread multi-criteria decision-making methods
is the Analytic Hierarchy Process (AHP). AHP successfully combines
the pairwise comparisons method and the hierarchical approach. It
allows the decision-maker to set priorities for all ranked alternatives.
But what if, for some of them, their ranking value is known (e.g.,
it can be determined differently)? The Heuristic Rating Estimation
(HRE) method proposed in 2014 tried to bring the answer to this question.
However, the considerations were limited to a model that did not consider
many criteria. In this work, we go a step further and analyze how
HRE can be used as part of the AHP hierarchical framework. The theoretical
considerations are accompanied by illustrative examples showing HRE
as a multiple-criteria decision-making method.
\end{abstract}
\begin{keyword}
pairwise comparisons \sep analytic hierarchy process \sep heuristic
ranking estimation \sep MCDM \sep AHP \sep HRE 
\end{keyword}

\end{frontmatter}{}

\section{Introduction}

The pairwise comparisons (PC) method has a centuries-old history.
The first considerations on its use come from a thirteenth-century
treatise by Ramon Lull, Ars Electionis \citep{Colomer2011rlfa,Haegele2001lwoe}.
Later, it was the subject of research by Condorcet, Fechner, Thurston
and others \citep{Condorcet1785esld,Dzhafarov2011tfi,Thurstone1994aloc,Ramik2020pcmt}.
Currently, pairwise comparisons are a valuable source of preferential
information in many decision-making methods, including TOPSIS \citep{Yuen2014cclo},
PROMETHEE \citep{Brans2016pm}, MACBETH \citep{BanaECosta2016otmf},
BWM \citep{Rezaei2015bwmc} or multiple-criteria sorting \citep{Kadzinski2015mabp}.
One of the best known methods of multiple-criteria decision making
based on pairwise comparisons is the Analytic Hierarchy Process (AHP)
\citep{Saaty2013dmwt,Kulakowski2020utahp}. The method defined by
Saaty in 1977 \citep{Saaty1977asmf} quickly gained popularity. It
became widely used in many areas of application, including performance
assessment of employees \citep{Lidinska2018amfp}, country competitiveness
\citep{Kramulova2016amfc}, supplier segmentation \citep{Rezaei2013mcss},
portfolio management \citep{Charouz201044amdm}, or Sustainability
\citep{Szabo2021aahp}. More application examples can be found in
the reviews \citep{Kou2016pcmi,Ho2018tsot,Darko2019roao}. It also
quickly gained opponents who very often rightly pointed out the imperfections
of the method \citep{BanaeCosta2008acao,Barzilai1994arrn,Dyer1990rota}.
One of the problems is the representation of measurable entities,
such as distance, weight or area \citep{Vasconcelos2019emem}. In
particular, if we can measure a given object and have an exact value,
there is no point in relying on experts' opinions on this entity.
However, even if we do not want to ask an expert to compare some alternatives,
it may turn out that the priorities after calculating the ranking
deviate from the ratios resulting from the measurements. The Heuristic
Rating Estimation (HRE) method is an attempt to deal with this problem.
The HRE was initially defined as two alternative methods (additive
and multiplicative) for calculating the ranking \citep{Kulakowski2014hrea,Kulakowski2015hreg}.
In this article, we extend this model to include a hierarchy that
allows multiple criteria for evaluating alternatives to be considered.
The HRE calculation procedure is based on the same concept as EVM
(Eigenvalue Method) \citep{Saaty1977asmf} and GMM (Geometric Mean
Method) \citep{Crawford1985anot}. That is, the priority of a given
alternative is equal to the weighted average value of the priorities
of the other alternatives \citep{Kulakowski2016srot}. This compatibility
allows the original HRE idea to be easily combined with these methods
within one hierarchical model (Example. \ref{exa:example-2}). Thus,
naturally, HRE becomes a multi-criteria method that complements AHP
in cases where some alternatives have measurable or previously known
ratings.

The presented work is a follow-up to \citep{Kulakowski2014hrea,Kulakowski2015hreg}.
The preliminary section introduces the reader to the issues related
to pairwise comparisons and priority deriving methods, such as EVM
and GMM. The HRE calculation procedure is described in (Sec. \ref{sec:Heuristic-Rating-Estimation}).
The next section (Sec. \ref{sec:Multiple-criteria-Heuristic-Rati})
fits HRE into a multi-criteria hierarchical model used in the context
of AHP. The theoretical description is supplemented by practical examples
showing the applications of HRE (Ex. \ref{exa:example-1}) and its
integration with AHP (Ex. \ref{exa:example-2}). 

\section{Preliminaries }

\subsection{Pairwise Comparisons\label{subsec:Pairwise-Comparisons}}

Let $A=\left\{ a_{1},\ldots,a_{n}\right\} $ be a set of alternatives
and $C=[c_{ij}]$ be a pairwise comparisons (PC) matrix, such that
$c_{ij}\in\mathbb{R}_{+}$ for $i,j\in\{1,..,n\}$. A single element
$c_{ij}$ corresponds to a direct comparison of the $i$-th and $j$-th
alternatives. A single entry $c_{ij}$ has a quantitative meaning.
For example, when an expert decides that the $a_{i}$ alternative
is twice more preferred than $a_{j}$, then $c_{ij}$ takes the value
$2$. As a result, the elements of the diagonal are ones, i.e. $C$
takes the form:

\[
C=\left[\begin{array}{cccc}
1 & c_{12} & \cdots & c_{1n}\\
c_{21} & 1 & \cdots & c_{2n}\\
\vdots & \vdots & \ddots & \vdots\\
c_{n1} & c_{n2} & \cdots & 1
\end{array}\right].
\]

Based on the information stored in $C$, a priority vector is calculated.
For the purpose of the article, we will denote it as $[w(a_{1}),\ldots,w(a_{n})]^{T}$
where $w$ is the function $w:A\rightarrow\mathbb{R_{_{+}}}$. We
say that $a_{i}$ is more preferred than $a_{j}$ (denoted as $a_{i}\succ a_{j}$)
if $w(a_{i})>w(a_{j})$. Similarly, $a_{i}$ is considered as equally
preferred as $a_{j}$ (denoted as $a_{i}\sim a_{j}$) when $w(a_{i})=w(a_{j})$.

In the ideal case $c_{ij}=w(a_{i})/w(a_{j})$ i.e. the results of
comparisons provided by experts perfectly match the computed ranking.
In practice, however, we may expect only that $c_{ij}\approx w(a_{i})/w(a_{j})$.
The reason for this inequality is data inconsistency, which is most
often the result of expert errors.
\begin{defn}
A PC matrix $C$ is called consistent if 
\begin{equation}
c_{ik}=c_{ij}c_{jk},\ \text{for every}\ i,j,k=1,\ldots,n\label{eq:consistent}
\end{equation}
\end{defn}
Every PC matrix is said to be inconsistent if it is not a consistent
PC matrix. 
\begin{defn}
A PC matrix $C$ is called reciprocal if
\begin{equation}
c_{ij}=\frac{1}{c_{ji}}\ \text{for every}\ i,j=1,...,n\label{eq:reciprocal}
\end{equation}
\end{defn}
Although the reciprocity condition is natural, the use of non-reciprocal
matrices can also be found in the literature \citep{Hovanov2008dwfg,Kulakowski2015ahre,Kulakowski2016srot}.
It is easily seen that every consistent PC matrix is also reciprocal,
but not reversely.

One of the essential issues in the PC methods concerns the question
of how the priority vector needs to be determined. The answer is not
straightforward because the expert's judgments may not be consistent\footnote{It can be shown that, for a consistent matrix, each ranking method
will lead to the same priority vector \citep[p.  96]{Kulakowski2020utahp}.}. There are many methods of priority vector estimation in the literature
\citep{Ishizaka2006htdp}. Two of the most popular prioritization
methods are the Eigenvector Method (EVM) \citep{Saaty1977asmf} and
the Geometric Mean Method (GMM) \citep{Crawford1985anot}.

In EVM, the priorities of alternatives are determined by the real
and positive solution of $Cw=\lambda_{\textit{max}}w_{\textit{max}}$,
where $\lambda_{\textit{max}}$ is the principal eigenvalue (spectral
radius) of $C$, and $w_{\textit{max}}$ is an appropriate eigenvector.
The vector of priorities $w$ is assumed to be $w_{\textit{max}}$
rescaled so that the sum of its components is $1$.

In GMM, the priority of an individual $a_{i}$ is calculated as the
geometric mean of the i-th row of $C$. The sum of components also
rescales the created ranking vector. Both EVM and GMM have been repeatedly
debated and analyzed, and both methods have their proponents and opponents
\citep{Barzilai1997dwfp,Saaty2003dmwta}.

Depending on how inconsistent the PC matrix is, calculating the priority
vector reflects the real preferences of respondents. To determine
the degree to which PC matrices are inconsistent, appropriate indices
are used. The most popular one comes from Saaty \citep{Saaty1977asmf}
and is defined as: 
\begin{equation}
\textit{CI}=\frac{\lambda_{max}-n}{n-1},\label{eq:CI}
\end{equation}
where $\lambda_{max}$ denotes the principal eigenvalue of the $n\times n$
matrix $C$. Another interesting inconsistency indicator comes from
Koczkodaj \citep{Koczkodaj1993ando}. For the $n\times n$ PC matrix
$C$, it is defined as:

\begin{equation}
\mathscr{K}(C)\overset{\textit{df}}{=}\max_{\substack{i,j,k\in\{1,...,n\}\\
i\neq j,\,j\neq k,\,k\neq i
}
}\left\{ \min\left\{ \left|1-\frac{c_{ij}}{c_{ik}c_{kj}}\right|,\left|1-\frac{c_{ik}c_{kj}}{c_{ij}}\right|\right\} \right\} .\label{eq:koczkodajII}
\end{equation}

In general, the more inconsistent the matrix $C$, the larger the
inconsistency indices. When the value of the inconsistency index is
considerable, the ranking results are not credible. Thus, in most
cases, it is essential to keep the inconsistency as small as possible.
More about different inconsistency indices can be found in \citep{Brunelli2018aaoi,Kulakowski2020iifi,Kazibudzki2016aeop}.

\subsection{Analytic Hierarchy Process}

A single PC matrix allows us to compare alternatives with respect
to only one criterion at a time. In practice, however, there is often
a need to take more than one criterion into account. For example,
when buying a car, we do not just choose a \textquotedbl better\textquotedbl{}
vehicle, but also pay attention to its price, terms of warranty, and
so on. A framework that allows the use of pairwise comparisons as
a multi-criteria method is provided by the Analytic Hierarchy Process
(AHP). The method was proposed by Saaty in 1977 \citep{Saaty1977asmf}
and it was one of the first\footnote{A similar method was proposed J. R. Miller in 1966 \citep{Miller1966taow}.}
to use PC matrices allowing for a large number of criteria. AHP is
able to handle multiple criteria by defining a hierarchy (a tree)
of alternatives and criteria. Hence, at the bottom of the hierarchical
tree defining the decision model, alternatives are compared with respect
to the lowest level criteria, then those criteria are compared against
each other with respect to the higher level of criteria, and so on.
Although the AHP hierarchy can be freely expanded, the most popular
models consist of three layers: alternatives $A=\{a_{1},...,a_{n}\}$,
criteria $S=\{s_{1},...,s_{m}\}$ and the root node of the hierarchy
called ``goal'' or ``purpose''.

As an example of a hierarchical AHP decision model, let us consider
the problem of selecting a company manager. Suppose there are three
candidates for this position: Andrew, Benjamin and Christopher. When
considering these alternatives, we take into account their experience
in managerial positions, education and interpersonal skills, as well
as the ability to work under pressure. We compare the candidates in
terms of these criteria. The one with the best results in our comparison
will become the CEO of the company. 

Going into details, according to the AHP method, we create a separate
PC matrix for each criterion. Thus, $C^{(s)}$ is a PC matrix corresponding
to the mutual comparisons of all three candidates with regard to the
criterion $s$. There are four criteria for assessing candidates:
experience $(\textit{ex})$, education $(\textit{ed})$, interpersonal
skills $(\textit{is})$, and stress resistance $(\textit{sr})$. Hence,
for example, $C^{(\textit{ex})}$ corresponds to the comparisons of
our candidates in terms of their experience. We assume that the following
four matrices were prepared as a result of the expert assessment.

\[
C^{(\textit{ex})}=\left[\begin{array}{ccc}
1 & 2 & 4\\
\frac{1}{2} & 1 & 2\\
\frac{1}{4} & \frac{1}{2} & 1
\end{array}\right],\textrm{ }C^{(\textit{ed})}=\left[\begin{array}{ccc}
1 & \frac{1}{2} & \frac{1}{8}\\
2 & 1 & \frac{1}{4}\\
8 & 4 & 1
\end{array}\right],
\]

\[
C^{(\textit{is})}=\left[\begin{array}{ccc}
1 & \frac{1}{2} & 3\\
2 & 1 & 6\\
\frac{1}{3} & \frac{1}{6} & 1
\end{array}\right],\ C^{(\textit{sr})}=\left[\begin{array}{ccc}
1 & \frac{1}{2} & 1\\
2 & 1 & 2\\
1 & \frac{1}{2} & 1
\end{array}\right]
\]
For each considered matrix $C^{(s)}$ we calculate\footnote{We use EVM \citep{Saaty1977asmf}.}
the priority vector. These are: 
\[
w^{(\textit{ex})}=\left[\frac{4}{7},\frac{2}{7},\frac{1}{7}\right]^{T},\ w^{(\textit{ed})}=\left[\frac{1}{11},\frac{2}{11},\frac{8}{11}\right]^{T},
\]
\[
w^{(\textit{is})}=\left[\frac{3}{10},\frac{6}{10},\frac{1}{10}\right]^{T},\ w^{(\textit{sr})}=\left[\frac{1}{4},\frac{2}{4},\frac{1}{4}\right]^{T}.
\]
When calculating the final priority vector, we must take into account
the fact that the criteria may contribute to a different extent to
achieving the goal. For this reason, we compare them in pairs. These
comparisons form one more $4\times4$ PC matrix $\widehat{C}=\left[\widehat{c}_{ij}\right]$
where $\widehat{c}_{ij}$ denotes the result of individual comparisons
between the criteria $s_{i}$ and $s_{j}$, where $s_{1},\ldots,s_{4}$
mean $(\textit{ex}),(\textit{ed}),(\textit{is}),$ and $(\textit{sr})$,
correspondingly. 
\[
\widehat{C}=\left[\begin{array}{cccc}
1 & 4 & 2 & 8\\
\frac{1}{4} & 1 & \frac{1}{2} & 2\\
\frac{1}{2} & 2 & 1 & 4\\
\frac{1}{8} & \frac{1}{2} & \frac{1}{4} & 1
\end{array}\right].
\]
The priority vector $\widehat{w}$ for $\widehat{C}$ is

\[
\widehat{w}=\left[\frac{8}{15},\frac{2}{15},\frac{4}{15},\frac{1}{15}\right]^{T}.
\]
Thus, according to the expert opinion, the first criterion; experience,
has the priority $8/15$, education $2/15$, interpersonal skills
4/15, and stress resistance -- $1/15$. The priorities of criteria
determine the degree to which the comparisons of alternatives affect
the overall score. Thus, they become weights scaling the results of
direct comparisons between alternatives (the final priority vector
is a linear combination of vectors $w^{(s)}$ where the scaling factors
come from $\widehat{w}$). The overall ranking $w$ is given as: 
\[
w=\stackrel[j=1]{m}{\sum}\hat{w}_{j}w^{(s_{j})}.
\]
In our case, we have: 
\[
w=\hat{w}_{1}w^{(ex)}+\hat{w}_{2}w^{(ed)}+\hat{w}_{3}w^{(ch)}+\hat{w}_{4}w^{(ag)}=
\]
\[
=\frac{8}{15}\left[\frac{4}{7},\frac{2}{7},\frac{1}{7}\right]^{T}+\frac{2}{15}\left[\frac{1}{11},\frac{2}{11},\frac{8}{11}\right]^{T}+\frac{4}{15}\left[\frac{3}{10},\frac{6}{10},\frac{1}{10}\right]^{T}+\frac{1}{15}\left[\frac{1}{4},\frac{2}{4},\frac{1}{4}\right]^{T}\approx
\]
\[
\approx\left[0.414,\thinspace0.370,\thinspace0.216\right]^{T}.
\]
The above priority vector translates to the observation that Andrew
is slightly more preferred to be a company manager than Benjamin,
and both are more preferred than Christopher.

\section{Heuristic Rating Estimation \label{sec:Heuristic-Rating-Estimation}}

\begin{figure}[h]
\begin{centering}
\includegraphics[width=0.7\textwidth]{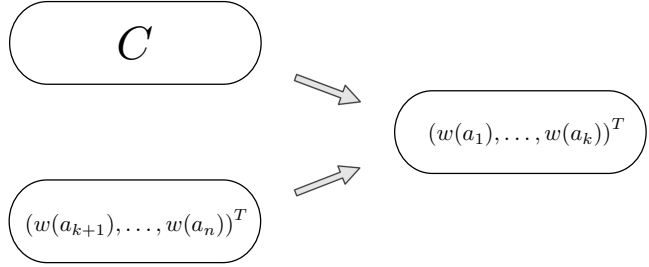}
\par\end{centering}
\centering{}\caption{How does the HRE priority deriving method work?}
\label{fig:hre-method-scheme}
\end{figure}

When comparing alternatives, sometimes, we already know priorities
for some of them. In such a case, we can use them and not ask experts
for unnecessary comparisons, saving time and money. This observation
gave rise to the Heuristic Rating Estimation (HRE) method \citep{Kulakowski2014hrea,Kulakowski2015hreg}.
In HRE, the set of alternatives $A$ is composed of two disjoint subsets:
$A_{K}$ - alternatives for which the final priorities are known,
and $A_{U}$ - alternatives for which priority weights need to be
estimated (also referred as to unknown alternatives). . Knowledge
about the priority values of elements in $A_{K}$ combined together
with pairwise comparisons of all the elements from $A=A_{K}\cup A_{U}$
allow us to calculate priorities for $A_{U}$. Hence, $A_{K}$ can
be considered as the set of references that provide a benchmark for
new alternatives in $A_{U}$. For the sake of simplicity, in this
study we will denote $A_{U}=\left\{ a_{1},...,a_{k}\right\} $ and
$A_{K}=\left\{ a_{k+1},...,a_{n}\right\} $. If $a_{i},a_{j}\in A_{K}$,
then we will assume that $w(a_{i})$ and $w(a_{j})$ are known. Therefore
the value $c_{ij}$ is also known and equals $c_{ij}=w(a_{i})/w(a_{j})$,
so querying experts about $c_{ij}$ is also not necessary. The comparisons
matrix $C$ in HRE takes the form: 
\[
C=\left[\begin{array}{cccccc}
1 & \cdots & c_{1,k} & c_{1k+1} & \cdots & c_{1,n}\\
\vdots & \ddots & \vdots & \vdots & \vdots & \vdots\\
c_{k,1} & \cdots & 1 & c_{k,k+1} & \cdots & c_{k,n}\\
c_{k+1,1} & \cdots & c_{k+1,k} & 1 & \cdots & w(a_{k+1})/w(a_{n})\\
\vdots & \cdots & \vdots & \vdots & \ddots & \vdots\\
c_{n,1} & \cdots & c_{n,k} & w(a_{n})/w(a_{k+1}) & \cdots & 1
\end{array}\right].
\]
The above matrix and vector of reference values $\left[w(a_{k+1}),...,w(a_{n})\right]^{T}$
allow us to calculate the complete ranking for $A$ (Fig. \ref{fig:hre-method-scheme}).

The HRE method has been designed as an extension of AHP where part
of the ranking is already known. Therefore, it is based on the same
premise that

\begin{equation}
w(a_{i})\approx c_{ij}w(a_{j}).\label{eq:observation-on-equiv}
\end{equation}
Thus, as in AHP, we can request that the priority of the i-th alternative
$w(a_{i})$ is a weighted average of priorities of all the other alternatives
\citep{Kulakowski2016srot}, i.e.

\begin{equation}
w(a_{i})=\frac{1}{n-1}\stackrel[\substack{j=1\\
j\neq i
}
]{n}{\sum}c_{ij}w(a_{j}),\,\,\,\text{for}\,\,\,i=1,\ldots,k.\label{eq:w_c_i HRE}
\end{equation}
This leads to a matrix equation

\begin{equation}
Mw=b,\label{eq:HRE}
\end{equation}
where $M$ is given as 

\begingroup\renewcommand*{\arraystretch}{1.2}

\[
M=\left[\begin{array}{cccc}
1 & -\frac{1}{n-1}c_{1,2} & \cdots & -\frac{1}{n-1}c_{1,k}\\
-\frac{1}{n-1}c_{2,1} & \ddots & \cdots & -\frac{1}{n-1}c_{2,k}\\
\vdots & \vdots & \ddots & \vdots\\
-\frac{1}{n-1}c_{k,1} & -\frac{1}{n-1}c_{k,2} & \cdots & 1
\end{array}\right],
\]
the constant term vector, 

\[
b=\left[\begin{array}{c}
\frac{1}{n-1}c_{1,k+1}w(a_{k+1})+...+\frac{1}{n-1}c_{1,n}w(a_{n})\\
\frac{1}{n-1}c_{2,k+1}w(a_{k+1})+...+\frac{1}{n-1}c_{2,n}w(a_{n})\\
\vdots\\
\frac{1}{n-1}c_{k,k+1}w(a_{k+1})+...+\frac{1}{n-1}c_{k,n}w(a_{n})
\end{array}\right]
\]

\endgroup

and the ranking vector for unknown alternatives is given as:
\[
w=\left[w(a_{1}),...,w(a_{k})\right]^{T}.
\]

The solution of (\ref{eq:HRE}) - if it exists and is acceptable,
forms the desired vector of priorities. It has been shown \citep{Kulakowski2015note}
that for a small value of $\mathfrak{\mathcal{\mathsf{\mathit{\mathfrak{\mathscr{K}}}}}}(C)$
and sufficiently large set of the reference alternatives $A_{K}$
the feasible solution of (\ref{eq:HRE}) always exists.

In HRE, as in AHP, many methods for calculating the ranking can be
defined. The GMM equivalent in AHP is geometric HRE \citep{Kulakowski2015hreg}.
Thus, starting from (\ref{eq:observation-on-equiv}) we may request

\begin{equation}
w(a_{i})=\left(\stackrel[j=1,j\neq i]{n}{\prod}c_{ij}w(a_{j})\right)^{\frac{1}{n-1}}\,\,\text{for}\,\,i=1,\ldots,n.\label{eq:w_c_i_HRE-g}
\end{equation}
 After raising both sides of (\ref{eq:w_c_i_HRE-g}) to the power
$n-1$, we obtain the non-linear equation system:

\begin{equation}
\begin{array}{c}
w^{n-1}(a_{1})=c_{1,2}w(a_{2})\cdot c_{1,3}w(a_{3})\cdot..........\cdot c_{1,n}w(a_{n})\\
w^{n-1}(a_{2})=c_{2,1}w(a_{1})\cdot c_{2,3}w(a_{3})\cdot..........\cdot c_{2,n}w(a_{n})\\
...............................................................................\\
w^{n-1}(a_{k})=c_{k,1}w(a_{1})\cdot c_{k,2}w(a_{2})\cdot...\cdot c_{k,n-1}w(a_{n-1})
\end{array}.\label{eq:uklad_HRE-g}
\end{equation}
Thanks to the logarithmic transformation, the above is equivalent
to the linear equation system

\begin{equation}
N\mu=d,\label{eq:HRE-g}
\end{equation}
where $N$ is a $k\times k$ auxiliary matrix in the form

\[
N=\left[\begin{array}{cccc}
(n-1) & -1 & \cdots & -1\\
-1 & \ddots & \cdots & -1\\
\vdots & \vdots & \ddots & \vdots\\
-1 & -1 & \cdots & (n-1)
\end{array}\right],
\]
$d$ is a constant term vector
\[
d=\left[\begin{array}{c}
\lg\left(c_{1,2}c_{1,3}\cdot...\cdot c_{1,k}c_{1,k+1}w(a_{k+1})\cdot...\cdot c_{1,n}w(a_{n})\right)\\
\lg\left(c_{2,1}c_{2,3}\cdot...\cdot c_{2,k}\cdot c_{2,k+1}w(a_{k+1})\cdot...\cdot c_{2,n}w(a_{n})\right)\\
\vdots\\
\lg\left(c_{k,1}c_{k,2}\cdot...\cdot c_{k,k-1}\cdot c_{k,k+1}w(a_{k+1})\cdot...\cdot c_{k,n}w(a_{n})\right)
\end{array}\right],
\]
and
\[
\mu=\left[\begin{array}{c}
\mu_{1}\\
\vdots\\
\vdots\\
\mu_{k}
\end{array}\right]=\left[\begin{array}{c}
\lg w(a_{1})\\
\vdots\\
\vdots\\
\lg w(a_{k})
\end{array}\right].
\]

The solution $\mu$ of (\ref{eq:HRE-g}) induces the weight vector
$w$ -- a solution of the original non-linear problem (\ref{eq:uklad_HRE-g})
\[
w=[\exp\mu_{1},...,\exp\mu_{k}]^{T}.
\]
The feasible solution of (\ref{eq:HRE-g}) always exists and is optimal
\citep{Kulakowski2015hreg}.

\section{Multiple-criteria Heuristic Rating Estimation Method\label{sec:Multiple-criteria-Heuristic-Rati}}

Sometimes when we want to compare a few objects or select several
items from many others, we follow different criteria with varying
degrees of importance. Moreover, we may have preliminary information
about specific things. For example, when choosing a school for a child,
parents pay attention to the level of education and the percentage
of pupils entering the next school, the educational system, location,
social group from which the pupils come, or additional activities
offered by the school. Specific information about some of the considered
schools can be known, although the set of known criteria may differ
in each school. As another example, consider a company planning to
expand its business by offering some new goods for sale in addition
to existing ones. Before introducing new products or services, it
is necessary to conduct market research and then select the most profitable
products/services. The company has data on the currently offered goods
(i.e., sales volume, profitability, and popularity), only the data
on the goods considered for introduction to the market are unknown.
During pairwise comparisons, the different criteria could be taken
into account.

\begin{figure}[h]
\begin{centering}
\includegraphics[width=0.9\textwidth]{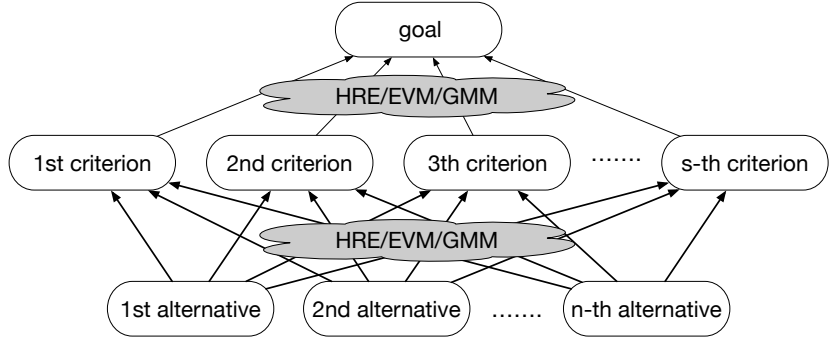}
\par\end{centering}
\centering{}\caption{The structure of hierarchical HRE}
\label{fig:hre-hierarchy-scheme}
\end{figure}

The cases described above have two distinctive features: some of the
alternatives considered have initially known priorities, and all the
alternatives are compared with respect to more than one criterion.
This observation leads us to the idea of the hierarchical HRE method.
Such an approach would reduce the multi-criteria problem to a series
of simple pairwise comparisons and calculate the weights of concepts
related to individual criteria. Finally, the priorities of the alternatives
are determined. Moreover, based on previously known estimations, we
would be able to estimate actual values of different decision options.
Hence, the meaning of the priority is not only relative but real,
expressing the current value or unit. In general, using a hierarchy
is nothing new. One of the first hierarchical methods\footnote{The concept of a hierarchy was taken by Saaty from Miller \citep{Miller1966taow}. }
- AHP - was introduced by Saaty and is still one of the most popular
decision-making methods. The structure of the hierarchical HRE is
shown in Fig. \ref{fig:hre-hierarchy-scheme}

The primary data for the hierarchical HRE method are two sets: a set
of criteria $Q=\{q_{1},..,q_{s}\}$ and a set of alternatives $A=\{a_{1},...,a_{n}\}$.
For each criterion $q_{t}$ $(t\in\left\{ 1,2,\ldots,s\right\} )$
we construct a PC matrix $C^{(q_{t})}=[c_{ij}^{(t)}]$ containing
comparisons of elements from $A$. For this purpose, we establish
a set of reference alternatives (with known priorities) $A_{K}^{(q_{t})}$,
and a set of unknown alternatives (i.e. with unknown priorities) $A_{U}^{(q_{t})}$.
Obviously, those sets for each $t\in\left\{ 1,2,\ldots,s\right\} $,
must satisfy the following conditions $A=A_{K}^{(q_{t})}\cup A_{U}^{(q_{t})},\:A_{K}^{(q_{t})}\cap A_{U}^{(q_{t})}=\slashed{O}$.
It is worth noting that the set of initially known alternatives and
set of unknown alternatives may differ among criteria.

We construct the PC matrix $C^{(q_{t})}$ based on pairwise comparisons
of alternatives, where at least one is unknown. Then we apply the
HRE prioritization method for the matrix $C^{(q_{t})}$. More specifically,
we construct the following linear equation:

\begin{equation}
M^{(q_{t})}w^{(q_{t})}=b^{(q_{t})},\label{eq:HHRE}
\end{equation}
where $w^{(q_{t})}$ is a priority vector of alternatives $a_{i}\in A_{U}^{(q_{t})}$,
$M^{(q_{t})}$ is a $k_{t}\times k_{t}$ matrix, where $k_{t}$ means
the number of known alternatives ($k_{t}=\left|A_{K}^{(q_{t})}\right|$)
in the form 
\[
M^{(q_{t})}=\left[\begin{array}{cccc}
1 & -\frac{1}{n-1}c_{1,2}^{(t)} & \cdots & -\frac{1}{n-1}c_{1,k_{t}}^{(t)}\\
-\frac{1}{n-1}c_{2,1}^{(t)} & \ddots & \cdots & -\frac{1}{n-1}c_{1,k_{t}}^{(t)}\\
\vdots & \vdots & \ddots & \vdots\\
-\frac{1}{n-1}c_{k_{t},1}^{(t)} & -\frac{1}{n-1}c_{k_{t},2}^{(t)} & \cdots & 1
\end{array}\right]
\]

\begingroup\renewcommand*{\arraystretch}{1.2}and the constant term vector is given as

\begin{equation}
b^{(q_{t})}=\left[\begin{array}{c}
\frac{1}{n-1}c_{1,k_{t}+1}^{(t)}w_{K}^{(q_{t})}(a_{k_{t}+1}^{(t)})+\ldots+\frac{1}{n-1}c_{1,n}^{(t)}w_{K}^{(q_{t})}(a_{n}^{(t)})\\
\frac{1}{n-1}c_{2,k_{t}+1}^{(t)}w_{K}^{(q_{t})}(a_{k_{t}+1}^{(t)})+\ldots+\frac{1}{n-1}c_{2,n}^{(t)}w_{K}^{(q_{t})}(a_{n}^{(t)})\\
\vdots\\
\frac{1}{n-1}c_{k_{_{t}},k_{t}+1}^{(t)}w_{K}^{(q_{t})}(a_{k_{t}+1}^{(t)})+\ldots+\frac{1}{n-1}c_{k_{t},n}^{(t)}w_{K}^{(q_{t})}(a_{n}^{(t)})
\end{array}\right],\label{eq:b_q}
\end{equation}

\endgroup

where $w_{K}^{(q_{t})}$ is a priority vector for alternatives from
$A_{K}^{(q_{t})}$. We calculate the priority vector $w^{(q_{t})}$
by solving the equation (\ref{eq:HHRE}).

In the geometric version of the HRE approach \citep{Kulakowski2015hreg}
for each criterion $q_{t}$ and the appropriate PC matrix $C^{(q_{t})}$
we construct the following linear equation

\begin{equation}
N^{(q_{t})}\mu^{(q_{t})}=d^{(q_{t})},\label{eq:HHRE-g}
\end{equation}
where 
\[
N^{(q_{t})}=\left[\begin{array}{cccc}
(n-1) & -1 & \cdots & -1\\
-1 & \ddots & \cdots & -1\\
\vdots & \vdots & \ddots & \vdots\\
-1 & -1 & \cdots & (n-1)
\end{array}\right]
\]
is a matrix of dimensions $k_{t}\times k_{t}$ , ($k_{t}=\left|A_{K}^{(q_{t})}\right|$)
and

\begingroup\renewcommand*{\arraystretch}{1.2}

\[
d^{(q_{t})}=\left[\begin{array}{c}
log\left(c_{1,2}^{(t)}c_{1,3}^{(t)}\cdot\ldots\cdot c_{1,k_{t}}^{(t)}c_{1,k_{t}+1}^{(t)}w(a_{k_{t}+1}^{(t)})\cdot\ldots\cdot c_{1,n}^{(t)}w(a_{n}^{(t)})\right)\\
log\left(c_{2,1}^{(t)}c_{2,3}^{(t)}\cdot\ldots\cdot c_{2,k_{t}}^{(t)}c_{2,k_{t}+1}^{(t)}w(a_{k_{t}+1}^{(t)})\cdot\ldots\cdot c_{2,n}^{(t)}w(a_{n}^{(t)})\right)\\
\vdots\\
log\left(c_{k_{t},1}^{(t)}c_{k_{t},2}^{(t)}\cdot\ldots\cdot c_{k_{t},k_{t}-1}^{(t)}c_{k_{t},k_{t}+1}^{(t)}w(a_{k_{t}+1}^{(t)})\cdot\ldots\cdot c_{k_{t},n}^{(t)}w(a_{n}^{(t)})\right)
\end{array}\right],
\]

\endgroup

and $\mu^{(q_{t})}$ is a logarithmized priority vector of alternatives
$a_{i}\in A_{U}^{(q_{t})}$. We compute the vector $\mu^{(q_{t})}$
by solving (\ref{eq:HHRE-g}) and then we calculate the vector $w_{U}^{(q_{t})}=\left(w_{U}^{(q_{t})}(a_{1}^{(t)}),\ldots,w_{U}^{(q_{t})}(a_{k_{t}}^{(t)})\right)$
using the dependency 
\begin{equation}
\mu^{(q_{t})}=\left(\log w_{U}^{(q_{t})}(a_{1}^{(t)}),\ldots,\log w_{U}^{(q_{t})}(a_{k_{t}}^{(t)})\right)\,\,\,\,\text{for every }a_{i}\in A_{U}^{(q_{t})}.\label{eq:log}
\end{equation}

Based on the priorities given in $w_{U}^{(q_{t})}$ we create an $n$
dimensional vector $w^{(q_{t})}$ being a catenation of a priori known
and just calculated priorities:

\[
w^{(q_{t})}=\left[w_{U}^{(q_{t})},w_{K}^{(q_{t})}\right]^{T}.
\]

\selectlanguage{english}%

\selectlanguage{american}%
Similarly as in AHP, when calculating the final priority vector, we
must take into account the fact that the criteria $q_{1},...,q_{s}$
may contribute at different extents to the goal. Thus we construct
a PC matrix $\widehat{C}$, in which every single entry $\widehat{c}_{kl}$
corresponds to the result of comparison $q_{k}$ against $q_{l}$.
Then (using the HRE approach) we calculate the priority vector

\[
\widehat{w}=\left[\widehat{w}(q_{1}),\ldots,\widehat{w}(q_{s})\right]^{T}.
\]
The final vector of weights is a linear combination of vectors $w^{(q_{t})}$
where scaling factors come from $\widehat{w}$, i.e. 
\begin{equation}
w=\stackrel[t=1]{s}{\sum}\widehat{w}(q_{t})w^{(q_{t})}.\label{eq:wagiHHRE}
\end{equation}
Note that if an additive approach is used, the existence of a solution
to each of the HRE equations (\ref{eq:HRE}) used must be verified.

\section{Numerical examples\label{sec:Numerical-examples}}
\begin{example}
\label{exa:example-1} A company, managed by Mr. Smith, runs a sports
facility consisting of a sports swimming pool, gym and fitness club.
The company is developing dynamically. Therefore Mr. Smith intends
to expand the facility with two further investments. He considers
a bowling alley, a professional massage parlor, a trampoline fitness
point, or a recreational pool. He would spend a similar amount of
money on each project. However, to increase profitability, Mr. Smith
intends to conduct a market survey first. Market research will be
carried out based on several criteria: $Q=\{pr$ - average monthly
income (profitability), $du$ -durability of the equipment (how many
months it will serve), and $pop$ - possible increase in the popularity
of a given form of spending free time (also as a result of advertising
campaigns)$\}$. A group of specialists will apply the HRE hierarchical
approach to the analysis of the obtained data.

Let $A=\{a_{1}$ - a bowling alley, $a_{2}$ - a professional massage
salon, $a_{3}$ - a trampoline fitness point, $a_{4}$ - a recreational
pool, $a_{5}$ - a sports pool, $a_{6}$ - gym, $a_{7}$ - fitness
club$\}.$ The set of reference alternatives is the same for each
criterion $A_{K}=\{a_{5},a_{6},a_{7}\}.$

The first considered criterion is profitability $(pr)$. For objects
from the reference set, the values $w^{(pr)}(a_{5})=20\,\,\text{thous.}$
, $w^{(pr)}(a_{6})=12\,\,\text{thous.}$, $w^{(pr)}(a_{7})=9\,\,\text{thous.}$
represent the average monthly income from the previous year. Considering
profitability, experts create a PC matrix $C^{(pr)}$. Each $c_{ij}$
in the matrix $C^{(pr)}$ corresponds to the relative profitability
of $a_{i}$ with respect to $a_{j}$. We calculate the values $c_{ij}$
for $i,j\in\{5,6,7\}$, $i\neq j$ as $\frac{w(a_{i})}{w(a_{j})}$,
therefore

\begingroup\renewcommand*{\arraystretch}{1.2}

\[
C^{(pr)}=\left[\begin{array}{ccccccc}
1 & \frac{2}{3} & 2 & \frac{1}{2} & \frac{1}{2} & 1 & \frac{3}{2}\\
\frac{3}{2} & 1 & 2 & \frac{2}{3} & \frac{1}{2} & 1 & 2\\
\frac{1}{2} & \frac{1}{2} & 1 & \frac{1}{3} & \frac{1}{4} & 1 & \frac{2}{3}\\
2 & \frac{3}{2} & 3 & 1 & \frac{2}{3} & \frac{3}{2} & 1\\
2 & 2 & 4 & \frac{2}{3} & 1 & \frac{20}{12} & \frac{20}{9}\\
1 & 1 & 1 & \frac{3}{2} & \frac{12}{20} & 1 & \frac{20}{9}\\
\frac{2}{3} & \frac{1}{2} & \frac{3}{2} & 1 & \frac{9}{20} & \frac{9}{20} & 1
\end{array}\right].
\]

In order to estimate the vector $w^{(pr)}$, we use the HRE method,
so according to (\ref{eq:HHRE}) we construct the matrix $M^{(pr)}$
and vector $b^{(pr)}$ 
\[
M^{(lan)}=\left[\begin{array}{cccc}
1 & -\frac{1}{n-1}c_{1,2} & -\frac{1}{n-1}c_{1,3} & -\frac{1}{n-1}c_{1,4}\\
-\frac{1}{n-1}c_{2,1} & 1 & -\frac{1}{n-1}c_{2,3} & -\frac{1}{n-1}c_{2,4}\\
-\frac{1}{n-1}c_{3,1} & -\frac{1}{n-1}c_{3,2} & 1 & -\frac{1}{n-1}c_{3,4}\\
-\frac{1}{n-1}c_{4,1} & -\frac{1}{n-1}c_{4,2} & -\frac{1}{n-1}c_{4,3} & 1
\end{array}\right]
\]
\[
b^{(pr)}=\left[\begin{array}{c}
\frac{1}{n-1}c_{1,5}w^{(pr)}(a_{5})+\frac{1}{n-1}c_{1,6}w^{(pr)}(a_{6})+\frac{1}{n-1}c_{1,7}w^{(pr)}(a_{7})\\
\frac{1}{n-1}c_{2,5}w^{(pr)}(a_{5})+\frac{1}{n-1}c_{2,6}w^{(pr)}(a_{6})+\frac{1}{n-1}c_{2,7}w^{(pr)}(a_{7})\\
\frac{1}{n-1}c_{3,5}w^{(pr)}(a_{5})+\frac{1}{n-1}c_{3,6}w^{(pr)}(a_{6})+\frac{1}{n-1}c_{3,7}w^{(pr)}(a_{7})\\
\frac{1}{n-1}c_{4,5}w^{(pr)}(a_{5})+\frac{1}{n-1}c_{4,6}w^{(pr)}(a_{6})+\frac{1}{n-1}c_{4,7}w^{(pr)}(a_{7})
\end{array}\right].
\]

\endgroup

The equation (\ref{eq:HHRE})$M^{(pr)}w^{(pr)}=b^{(pr)}$, written
numerically, takes the form 
\[
\left[\begin{array}{cccc}
1 & -0.111 & -0.333 & -0.083\\
-0.25 & 1 & -0.333 & -0.111\\
-0.083 & -0.083 & 1 & -0.056\\
-0.333 & -0.25 & -0.5 & 1
\end{array}\right]\left[\begin{array}{c}
w^{(pr)}(a_{1})\\
w^{(pr)}(a_{2})\\
w^{(pr)}(a_{3})\\
w^{(pr)}(a_{4})
\end{array}\right]=\left[\begin{array}{c}
5.917\\
6.667\\
3.833\\
6.722
\end{array}\right]
\]
After solving the equation, we get the following weights $w^{(pr)}(a_{1})=11.164,$
$w^{(pr)}(a_{2})=13.667,$ $w^{(pr)}(a_{3})=6.863,$ $w^{(pr)}(a_{4})=17.292,$
the values of which represent the estimated average monthly income
for the objects $a_{1},a_{2},a_{3},a_{4}$. After normalization, the
resulting weights are:
\[
w^{(pr)}=[\begin{array}{ccccccc}
0.124 & 0.152 & 0.076 & 0.192 & 0.222 & 0.133 & 0.1\end{array}]^{T}.
\]

According to the second considered criterion $(du)$, we compare objects
in terms of the durability of the equipment. The values $c_{ij}$
in the matrix $C^{(du)}$ indicate how many times the durability of
the equipment of the object $a_{i}$ is better than the durability
of the equipment of the object $a_{j}.$ For reference alternatives,
the values $w^{(du)}(a_{5})=72$, $w^{(du)}(a_{6})=24$, $w^{(du)}(a_{7})=36$
represent the average number of months the appliance has been used
in the last few years. The PC matrix $C^{(du)}$ prepared by the experts
has the form

\begingroup\renewcommand*{\arraystretch}{1.2}

\[
C^{(du)}=\left[\begin{array}{ccccccc}
1 & 3 & 2 & 1 & \frac{1}{2} & 2 & \frac{3}{2}\\
\frac{1}{3} & 1 & \frac{3}{2} & \frac{1}{3} & \frac{1}{4} & \frac{1}{2} & \frac{1}{2}\\
\frac{1}{2} & \frac{2}{3} & 1 & \frac{1}{3} & \frac{1}{4} & \frac{2}{3} & \frac{1}{2}\\
1 & 3 & 3 & 1 & \frac{4}{5} & 2 & 2\\
2 & 4 & 4 & \frac{5}{4} & 1 & \frac{72}{24} & \frac{72}{36}\\
\frac{1}{2} & 2 & \frac{3}{2} & \frac{1}{2} & \frac{24}{72} & 1 & \frac{24}{36}\\
\frac{2}{3} & 2 & 2 & \frac{1}{2} & \frac{36}{72} & \frac{36}{24} & 1
\end{array}\right].
\]

\endgroup

Using HRE for the matrix $C^{(du)}$, similarly to the first criterion
(see (\ref{eq:HHRE}) and (\ref{eq:b_q})), we get the matrix $M^{(du)}$
and vector $b^{(du)}$: 
\[
M^{(du)}=\left[\begin{array}{cccc}
1 & -0.5 & -0.333 & -0.167\\
-0.056 & 1 & -0.25 & -0.056\\
-0.083 & -0.111 & 1 & -0.056\\
-0.167 & -0.5 & -0.5 & 1
\end{array}\right],
\]
\[
b^{(du)}=\left[\begin{array}{c}
23\\
8\\
8.667\\
29.6
\end{array}\right].
\]
Solving the equation $M^{(du)}w^{(du)}=b^{(du)}$, gives values that
indicate the estimated average durability in months
\[
w^{(du)}(a_{1})=47.183,\;w^{(du)}(a_{2})=18.119,\;w^{(du)}(a_{3})=17.688,\;w^{(du)}(a_{4})=55.367,
\]
Scaling the solution vector gives:
\[
w^{(du)}=[\begin{array}{ccccccc}
0.174 & 0.067 & 0.065 & 0.205 & 0.266 & 0.089 & 0.133\end{array}]^{T}.
\]

Analysis of the third criterion -- the possibility of increasing
the popularity of activities offered by individual facilities $(pop)$
-- the values $c_{ij}$ in the matrix $C^{(pop)}$ mean how many
times the popularity of the object $a_{i}$ can increase in relation
to the increase in the popularity of the object $a_{j}$. The reference
alternatives have the values $w^{(pop)}(a_{5})=5$, $w^{(pop)}(a_{6})=20$,
$w^{(pop)}(a_{7})=25$, and represent the average percentage increase
in popularity in recent years. In this case, the PC - matrix $C^{(pop)}$
proposed by the experts is

\begingroup\renewcommand*{\arraystretch}{1.2}

\[
C^{(pop)}=\left[\begin{array}{ccccccc}
1 & 1 & \frac{1}{5} & 3 & 3 & 2 & 2\\
1 & 1 & \frac{1}{3} & 3 & 3 & 2 & 2\\
5 & 3 & 1 & 4 & 5 & 3 & 3\\
\frac{1}{3} & \frac{1}{3} & \frac{1}{4} & 1 & \frac{3}{2} & \frac{1}{2} & \frac{1}{2}\\
\frac{1}{3} & \frac{1}{3} & \frac{1}{5} & \frac{2}{3} & 1 & \frac{5}{20} & \frac{5}{25}\\
\frac{1}{2} & \frac{1}{2} & \frac{1}{3} & 2 & \frac{20}{5} & 1 & \frac{20}{25}\\
\frac{1}{2} & \frac{1}{2} & \frac{1}{3} & 2 & \frac{25}{5} & \frac{25}{20} & 1
\end{array}\right].
\]

\endgroup

The auxiliary matrix $M^{(pop)}$ and the vector $b^{(pop)}$ derived
from the matrix $C^{(pop)}$ are
\[
M^{(pop)}=\left[\begin{array}{cccc}
1 & -0.167 & -0.033 & -0.5\\
-0.167 & 1 & -0.056 & -0.5\\
-0.833 & -0.5 & 1 & -0.667\\
-0.056 & -0.056 & -0.042 & 1
\end{array}\right],
\]
\[
b^{(pop)}=\left[\begin{array}{c}
17.5\\
17.5\\
26.667\\
5
\end{array}\right].
\]
Using the HRE method to solve the equation $M^{(pop)}w^{(pop)}=b^{(pop)}$,
we obtain the following values:
\[
w^{(pop)}(a_{1})=31.459,\;w^{(pop)}(a_{2})=32.93,\;w^{(pop)}(a_{3})=77.21,\;w^{(pop)}(a_{4})=11.794,
\]
These numbers represent the estimated average growth in popularity
of objects $a_{1},a_{2},a_{3},a_{4}$. After normalization, the solution
vector of the weights is equal:
\[
w^{(du)}=[\begin{array}{ccccccc}
0.155 & 0.162 & 0.38 & 0.058 & 0.025 & 0.098 & 0.123\end{array}]^{T}.
\]
Since in the considered case the weights of the criteria are known
and equal
\[
\hat{w}=[\begin{array}{ccc}
0.5 & 0.2 & 0.3\end{array}]^{T}.
\]
the final priorities of individual objects are calculated according
to the schema (\ref{eq:wagiHHRE}):
\[
w(a_{i})=0.5\cdot w^{(pr)}(a_{i})+0.2\cdot w^{(du)}(a_{i})+0.3\cdot w^{(pop)}(a_{i}).
\]
So the final weight vector is equal
\[
w=[\begin{array}{ccccccc}
0.143 & 0.138 & 0.165 & 0.154 & 0.172 & 0.114 & 0.114\end{array}]^{T},
\]
which means that investing in a trampoline fitness point and a recreational
pool may be slightly more profitable than any of the other two new
facilities.
\end{example}

\medskip{}

\begin{example}
\label{exa:example-2}Mr. Kowalski became interested in collecting
porcelain. He paid particular attention to cups. Mr. Kowalski visited
a local antique store and found five exciting items. He would like
to buy the two of them that are the most valuable. Unfortunately,
there is no universal algorithm that allows him to evaluate porcelain
step by step. Moreover, the valuation of porcelain is complex and
immeasurable due to the specificity of various markets. The identical
cups may have different prices depending on the region of the world
in which they are offered. For example, Polish collectors are very
fond of German products, undervaluing Polish porcelain a bit. At the
same time, the English are proud of their products and are eager to
buy them; they also value Chinese porcelain. Therefore, when analyzing
trends in the porcelain market, we should also remember the purchase
location and ask an expert to evaluate various criteria according
to which they make a purchase decision. Let us assume that Mr. Kowalski
lives in Poland, so we adopt the values of individual criteria according
to the Polish market and will help him make the right decision. Our
market research shows that the final value of the cups depends here
on several factors.

The most straightforward and most obvious rule is that the more difficult
it is to get products from a specific series, period, or manufacturer,
the higher the price it receives. Due to the fragility of porcelain,
whole sets are more expensive than single pieces. Some units have
special status: they are almost unattainable on the market, they are
valued as \textquotedbl unlimited\textquotedbl{} and are usually
auctioned at exorbitant prices. The products from the first period
of operation of the factory in Meissen, Germany, run by the \textquotedbl father
of European porcelain,\textquotedbl{} Friedrich B?tger, are a good
example. Moreover, products of a particular manufacturer may be more
in demand at one time and less at another.

The second criterion is the reputation of the manufacturer. It is
closely influenced by the history of a given factory, the quality
of the porcelain produced there, or its uniqueness. It's also worth
remembering that reputation is tied to the current fashion - it depends
not only on time but also on location. For example: in Germany, Meissen
porcelain is the most valued, while in Poland, collectors like Rosenthal
more.

Traces, called signatures, can often be found on porcelain. They define
its origin, age, model name, and sometimes the project's author. Some
signatures even specify the type of porcelain the item is made of.
Thanks to the signature, we can estimate the value of the product.
Usually, the older it is, the more valuable it is. The first characters
were discovered on Chinese porcelain from the 14th century. The well-known
collector's rarities include, among others, unique items from the
Ming Dynasty. There are no more than a dozen of them in museum collections
worldwide. They are almost nonexistent at auctions and, if they do
appear, they reach exorbitant prices. In Europe, the first porcelain
products were made in Saxony in the 18th century during the reign
of Augustus II and were marked with letters and numbers. Soon, other
manufacturers started putting their signatures on the bottom of dishes.
Depending on the time of creation, the signatures differ in color
and how they were made (painting, printing, decal). The markings also
changed within one factory, for example, with a change of ownership.
On their basis, the experienced collector can accurately determine
the creation time.

Ancient porcelain is extremely fragile, which, of course, increases
its value. However, in the event of damage - even porcelain from an
excellent manufacturer is worthless. Therefore, before buying, you
should evaluate the condition of a single piece of porcelain. An ideal
copy will be a product that does not have any acquired or primary
defects. By primary defects, we mean any flaws visible in the product
that are traces of an ineffective porcelain forming process - stains,
painting, or decals errors. Reputable manufacturers sold only first-class
products. For example, all defective products are considered broken
at Rosenthal factories, so they do not end up on the market. Interestingly,
primary defects usually do not influence the product's price, unlike
acquired defects. Acquired defects affecting porcelain price include
nicks, sticking together, painting with other paint, or visible spider
web shape cracks.

The quality of porcelain is primarily determined by the appropriate
selection of raw materials, especially a large amount of kaolin. Good
porcelain is hard, durable, and scratch-resistant. It requires manual
work as machines do not provide a top-quality finish. Therefore, each
item is processed precisely and accurately by hand in good factories.
This guarantees the highest quality and timeless beauty of the final
product. Ornaments on porcelain also determine its quality. The most
valuable is hand-decorated porcelain. Often, ornaments are made of
noble materials - real gold or, for example, platinum.

Based on the above information, we will compare the value of considered
porcelain cups against five criteria: factory, period of creation,
uniqueness, state of preservation and quality. Let

\[
Q=\{\textit{man},\textit{per},\textit{un},\textit{st},\textit{qua}\}
\]

denote a set of significant criteria, where $\textit{man},\textit{per},\textit{un},\textit{st},\textit{qua}$
mean factory, period of creation, uniqueness, state of preservation
and quality, respectively.

Let $A=\{$$a_{1},a_{2}$,$a_{3}$,$a_{4},a_{5}\}$ denote a considered
collection of cups. First, we compare the cups in terms of the first
criterion. Depending on the reputation of the factory, each cup is
assigned a value from 1 to 10, where 1 denotes the lowest possible
score.  All cups come from different factories. We know three of
them (for the alternatives $a_{3}$, $a_{4}$ and $a_{5}$) and we
can assign values to them based on our experience. They are $w^{(\textit{man})}(a_{3})=8.7,$
$w^{(\textit{man})}(a_{4})=4.2,$ $w^{(\textit{man})}(a_{5})=7.2$
respectively. As it is the first time that we are dealing with the
products of manufacturers from which the cups $a_{1}$ and $a_{2}$
come from, we will apply the HRE procedure. As a result, we get the
matrix:

\begingroup\renewcommand*{\arraystretch}{1.2}

\[
C^{(man)}=\left[\begin{array}{ccccc}
1 & 2 & \frac{1}{3} & 3 & 1\\
\frac{1}{2} & 1 & \frac{1}{5} & 1 & \frac{1}{2}\\
3 & 5 & 1 & \frac{8.7}{4.2} & \frac{8.7}{7.2}\\
\frac{1}{3} & 1 & \frac{4.2}{8.7} & 1 & \frac{4.2}{7.2}\\
1 & 2 & \frac{7.2}{8.7} & \frac{7.2}{4.2} & 1
\end{array}\right].
\]

\endgroup

Since the weight of cups $a_{3},\:a_{4}$ and $a_{5}$ are known,
the pairs $(a_{3},a_{4})$, $(a_{3},a_{5})$ and $(a_{4},a_{5})$
are not evaluated, but $c_{ij}^{(man)}$ is calculated as $\frac{w^{(\textit{man})}(i)}{w^{(\textit{man})}(j)}$
$(3\leq i<j\leq5)$. To estimate the expected values of factories
A and B we follow the HRE procedure. Since $A_{U}^{((\textit{man})}=\{a_{1},a_{2}\}$
and $A_{K}^{(\textit{man})}=\{a_{3},a_{4},a_{5}\}$, the matrix $M^{(\textit{man})}$
and the constant term vector $b^{(\textit{man})}$ have the form (see
\ref{eq:HHRE}) 
\[
M^{(\textit{man})}=\left[\begin{array}{cc}
1 & -\frac{1}{n-1}c_{1,2}\\
-\frac{1}{n-1}c_{2,1} & 1
\end{array}\right],
\]
\[
b^{(man)}=\left[\begin{array}{c}
\frac{1}{n-1}c_{1,3}w^{(\textit{man})}(a_{3})+\frac{1}{n-1}c_{1,4}w^{(\textit{man})}(a_{4})+\frac{1}{n-1}c_{1,5}w^{(\textit{man})}(a_{5})\\
\frac{1}{n-1}c_{2,3}w^{(\textit{man})}(a_{3})+\frac{1}{n-1}c_{2,4}w^{(\textit{man})}(a_{4})+\frac{1}{n-1}c_{2,5}w^{(\textit{man})}(a_{5})
\end{array}\right].
\]
 Hence, numerically:
\[
M^{(\textit{man})}=\left(\begin{array}{cc}
1 & -0.476\\
-0.119 & 1
\end{array}\right),\;b^{(man)}=\left[\begin{array}{c}
5.675\\
2.385
\end{array}\right].
\]
As $M^{(\textit{man})}w^{(\textit{man})}=b^{(\textit{man})}$ has
an admissible solution (i.e. real and positive), the ranking vector
$w^{(man)}$ gets the form
\[
w^{(man)}=[\begin{array}{ccccc}
7.22 & 3.244 & 8.7 & 4.2 & 7.2\end{array}]^{T}.
\]
 Hence, according to experts' judgments $w^{(man)}(a_{1})=7.22$ and
$w^{(man)}(a_{2})=3.244$. After rescaling 
\[
w^{(man)}=[\begin{array}{ccccc}
0.236 & 0.106 & 0.284 & 0.137 & 0.235\end{array}]^{T}.
\]

The second considered criterion applies to the date the cup was manufactured.
Based on the signatures placed on the cups $a_{1}$, $a_{3}$ and
$a_{5}$ we can read the years of production and assign the appropriate
number of points according to age. The older the cup, the more points
it gets. Finally, we receive the results $w^{(\textit{per})}(c_{1})=8,$
$w^{(\textit{per})}(c_{3})=10,$ $w^{(\textit{per})}(c_{5})=5$. The
cup $a_{2}$ has no signature and the signature of the cup $a_{4}$
does not include the year, so in order to establish the period of
creation for these two cups we ask an expert for help. In the case
of this criterion $A_{U}^{(\textit{per})}=\{a_{2},a_{4}\}$ and $A_{K}^{(\textit{per})}=\{a_{1},a_{3},a_{5}\}$.
As a result of the expert's assessment, we obtain the matrix:

\begingroup\renewcommand*{\arraystretch}{1.2}

\[
C^{(per)}=\left[\begin{array}{ccccc}
1 & 2 & \frac{8}{10} & 4 & \frac{8}{5}\\
\frac{1}{2} & 1 & \frac{1}{3} & 2 & 1\\
\frac{10}{8} & 3 & 1 & 5 & \frac{10}{5}\\
\frac{1}{4} & \frac{1}{2} & \frac{1}{5} & 1 & \frac{1}{2}\\
\frac{5}{8} & 1 & \frac{5}{10} & 2 & 1
\end{array}\right].
\]

\endgroup

In order to obtain the values of the priority vector $w^{(\textit{per})}$
for cups $a_{2}$ and $a_{4}$ it is enough to create the matrix $M^{(\textit{per})}$
and vector $b^{(per)}$ using the HRE method and then solve the equation
$M^{(\textit{per})}w^{(\textit{per})}=b^{(\textit{per})}$. The matrix
$M^{(\textit{per})}$ can be obtained from the matrix $C^{(\textit{per})}$
by removing the columns and rows corresponding to known alternatives
from the set $A_{K}^{(\textit{per})}$ and multiplying values out
of a diagonal by $-\frac{1}{n-1}$. The values removed only from the
columns (but not from the rows) are used to form the vector $b^{(\textit{per})}$.
Thus
\[
M^{(\textit{per})}=\left[\begin{array}{cc}
1 & -\frac{1}{n-1}c_{2,4}\\
-\frac{1}{n-1}c_{4,2} & 1
\end{array}\right],
\]
\[
b^{(\textit{per})}=\left[\begin{array}{c}
\frac{1}{n-1}c_{2,1}w^{(\textit{per})}(a_{1})+\frac{1}{n-1}c_{2,3}w^{(\textit{per})}(a_{3})+\frac{1}{n-1}c_{2,5}w^{(\textit{per})}(a_{5})\\
\frac{1}{n-1}c_{4,1}w^{(\textit{per})}(a_{1})+\frac{1}{n-1}c_{4,3}w^{(\textit{per})}(a_{3})+\frac{1}{n-1}c_{4,5}w^{(\textit{per})}(a_{5})
\end{array}\right],
\]
 and the equation $M^{(\textit{per})}w^{(\textit{per})}=b^{(\textit{per})}$
takes the shape
\[
\left(\begin{array}{cc}
1 & -0.498\\
-0.124 & 1
\end{array}\right)\left[\begin{array}{c}
w^{(\textit{per})}(a_{2})\\
w^{(\textit{per})}(a_{4})
\end{array}\right]=\left(\begin{array}{c}
\frac{37}{12}\\
\frac{13}{8}
\end{array}\right).
\]
The solution of the above equation is the following vector 
\[
w^{(\textit{per})}=[\begin{array}{ccccc}
8 & 4.15 & 10 & 2.14 & 5\end{array}]^{T}.
\]
 Thus, in the category ``period of creation'' the expected number
of points on a scale from 1 to 10 for cups $a_{2}$ and $a_{4}$ are
$4.15$ and $2.14$ respectively. After normalization

\[
w^{(per)}=[\begin{array}{ccccc}
0.273 & 0.141 & 0.341 & 0.073 & 0.171\end{array}]^{T}.
\]

The third considered criterion concerns uniqueness, although this
time the set of reference alternatives is empty. Thus, once again,
we ask an expert to compare in pairs the profitability of purchasing
a given cup in terms of its uniqueness on the market. His assessments
form the following PC matrix

\begingroup\renewcommand*{\arraystretch}{1.2}

\[
C^{(un)}=\left[\begin{array}{ccccc}
1 & \frac{1}{2} & \frac{1}{2} & \frac{1}{2} & \frac{1}{3}\\
2 & 1 & 2 & 1 & \frac{1}{2}\\
2 & \frac{1}{2} & 1 & 1 & \frac{1}{2}\\
2 & 1 & 1 & 1 & \frac{1}{2}\\
3 & 2 & 2 & 2 & 1
\end{array}\right].
\]

\endgroup3

In such a case, we can use EVM (Sec. \ref{subsec:Pairwise-Comparisons})
i.e. solve the equation $C^{(\textit{un})}w=\lambda_{\textit{max}}w$.
Hence, after appropriate rescaling of the vector $w$ we get

\[
w^{(un)}=[\begin{array}{ccccc}
0.097 & 0.214 & 0.161 & 0.182 & 0.345\end{array}]^{T}.
\]

The condition of the cup depends mainly on two factors: primary damage
and acquired damage. Therefore, the experts assess all our alternatives
against two sub-criteria: level of primary damage and level of acquired
damage. As a result of comparing the degree of primary damage, we
obtain the following matrix:

\begingroup\renewcommand*{\arraystretch}{1.2}

\[
C^{(\textit{pd})}=\left(\begin{array}{ccccc}
1 & 9 & 9 & \frac{8}{3} & \frac{5}{3}\\
\frac{1}{9} & 1 & \frac{4}{9} & \frac{1}{9} & \frac{1}{8}\\
\frac{1}{9} & \frac{9}{4} & 1 & \frac{1}{9} & \frac{1}{8}\\
\frac{3}{8} & 9 & 9 & 1 & 2\\
\frac{3}{5} & 8 & 8 & \frac{1}{2} & 1
\end{array}\right).
\]

The comparisons of acquired damage are as follows:

\[
C^{(\textit{ad})}=\left(\begin{array}{ccccc}
1 & \frac{1}{9} & \frac{1}{9} & \frac{4}{5} & \frac{1}{9}\\
9 & 1 & \frac{3}{4} & 9 & \frac{6}{5}\\
9 & \frac{4}{3} & 1 & 9 & \frac{9}{4}\\
\frac{5}{4} & \frac{1}{9} & \frac{1}{9} & 1 & \frac{1}{6}\\
9 & \frac{5}{6} & \frac{4}{9} & 6 & 1
\end{array}\right).
\]

\endgroup

The priority vectors for the $C^{(\textit{pd})}$ and $C^{(\textit{ad})}$
matrices calculated using EVM are as follows:

\[
w^{(\textit{pd})}=\left[\begin{array}{ccccc}
0.416 & 0.029 & 0.04 & 0.289 & 0.223\end{array}\right]^{T},
\]

and 

\[
w^{(\textit{ad})}=\left[\begin{array}{ccccc}
0.033 & 0.301 & 0.39 & 0.038 & 0.235\end{array}\right]^{T}.
\]

Comparing the significance of both types of damage, the expert found
that the acquired damage is two times more important than the primary
damage. Hence, the ranking of both sub-criteria is as follows:

\[
w^{(\textit{st-c})}=\left[\begin{array}{c}
w^{(\textit{st-c})}(\textit{pd})\\
w^{(\textit{st-c})}(\textit{ad})
\end{array}\right]=\left[\begin{array}{c}
0.333\\
0.666
\end{array}\right].
\]
The state of preservation for the i-th cup is calculated as 
\[
w^{(\textit{st})}(a_{i})=w^{(\textit{st-c})}(\textit{pd})\cdot w^{(\textit{pd})}(a_{i})+w^{(\textit{st-c})}(\textit{ad})\cdot w^{(\textit{ad})}(a_{i}),
\]
for $i=1,\ldots,5$, hence, finally the assessment of the state of
the cups' preservation is as follows:

\[
w^{(\textit{st})}=\left[\begin{array}{ccccc}
0.16 & 0.211 & 0.27 & 0.12 & 0.23\end{array}\right]^{T}.
\]

When assessing the quality of porcelain, the expert considered its
resistance to scratching. In the case of two cups $a_{1}$ and $a_{2}$,
it was known and calculated as the average number of significant scratches
among a similar class of products appearing at online auctions in
the last five years. Thus, $A_{U}^{(\textit{qua})}=\{a_{3},a_{4},a_{5}\}$
and $A_{K}^{(\textit{qua})}=\{a_{1},a_{2}\}$, where $w^{(\textit{qua})}(a_{1})=5.7$
and $w^{(\textit{qua})}(a_{2})=2.4$. As a result of comparing the
quality of alternatives by an expert, we get the matrix

\begingroup\renewcommand*{\arraystretch}{1.2}
\[
C^{(\textit{qua})}=\left(\begin{array}{ccccc}
1 & \frac{5.7}{2.4} & 9 & 3 & \frac{7}{5}\\
\frac{2.4}{5.7} & 1 & 9 & \frac{9}{5} & \frac{8}{3}\\
\frac{1}{9} & \frac{1}{9} & 1 & \frac{1}{9} & \frac{1}{9}\\
\frac{1}{3} & \frac{5}{9} & 9 & 1 & \frac{6}{5}\\
\frac{5}{7} & \frac{3}{8} & 9 & \frac{5}{6} & 1
\end{array}\right).
\]

\endgroup

Based on the HRE approach, the auxiliary matrix and the constant term
vector are 
\[
M^{(\textit{qua})}=\left(\begin{array}{ccc}
1 & -0.0259 & -0.0259\\
-2.104 & 1 & -0.28\\
-2.104 & -0.194 & 1
\end{array}\right),\,\,\,b^{(\textit{qua})}=\left(\begin{array}{c}
0.225\\
0.808\\
1.242
\end{array}\right).
\]

Solving the HRE equation $M^{(\textit{qua})}w=b^{(\textit{qua})}$
reveals that the expected number of scratches for $a_{3},a_{4}$ and
$a_{5}$ are $0.344,2.2$ and $2.39$. This allows us to form the
ranking vector

\[
\widehat{w}^{(\textit{qua})}=[\begin{array}{ccccc}
5.7 & 2.4 & 0.344 & 2.2 & 2.39\end{array}]^{T}.
\]

However, since the ranking values represent the expected number of
scratches, it means that the higher the value, the worse the product.
To change this, we will raise each of the elements of $\widehat{w}^{(\textit{qua})}$
to the -1 power and then normalize it. As a result, we get the vector:

\[
w^{(\textit{qua})}=[\begin{array}{ccccc}
0.04 & 0.0954 & 0.665 & 0.104 & 0.0955\end{array}]^{T}.
\]
It is worth noting that, thanks to the last transformation, if the
i-th alternative is two times worse than the j-th one according to
$\widehat{w}^{(\textit{qua})}$ then according to $w^{(\textit{qua})}$
it is two times better than the j-th alternative etc. The obtained
vector is a ranking of the cups against the porcelain quality criterion.

In order to establish the weights of criteria ($q_{1}=man,\:q_{2}=per,\:q_{3}=un,\:q_{4}=st,\:q_{5}=qua$)
the expert constructed a PC-matrix, in which $\widehat{c}_{ij}$ denotes
how many times the criterion $q_{i}$ is more preferred than the criterion
$q_{j}$. 
\[
\widehat{C}=\left[\begin{array}{ccccc}
1 & 1 & \frac{3}{2} & 3 & 3\\
1 & 1 & \frac{3}{2} & 3 & 3\\
\frac{2}{3} & \frac{2}{3} & 1 & 2 & 2\\
\frac{1}{3} & \frac{1}{3} & \frac{1}{2} & 1 & 1\\
\frac{1}{3} & \frac{1}{3} & \frac{1}{2} & 1 & 1
\end{array}\right]
\]
Then, using EVM again, we get a priority vector for the criteria:
\[
\hat{w}=[\begin{array}{ccccc}
0.3 & 0.3 & 0.2 & 0.1 & 0.1\end{array}]^{T}.
\]
The final weights of individual cups $a_{1},...,a_{5}$ are calculated
according to the well-known schema:
\[
w(a_{i})=0.3w^{(man)}(a_{i})+0.3w^{(per)}(a_{i})+0.2w^{(un)}(a_{i})+0.1w^{(st)}(a_{i})+0.1w^{(qua)}(a_{i}).
\]
Thus we obtain
\[
w=[\begin{array}{ccccc}
0.1922 & 0.1478 & 0.3138 & 0.1222 & 0.2237\end{array}]^{T}.
\]
The recommendation presented to Mr. Kowalski puts in the first place
the cup $a_{3}$ with the rank $0.3138$, followed by $a_{5}$ with
the rank $0.2237$, then $a_{1},a_{2}$ and $a_{4}$ with the ranking
values $0.1922,\,0.1478$ and $0.1222$.
\end{example}
\medskip{}

\section{Summary}

In the work, we show the hierarchical HRE in the context of a multiple-criteria
decision-making process. This allows us to look at HRE not only as
a procedure for estimating the rating, but as a multiple-criteria
decision-making method. The concept of HRE is based on the same assumptions
that can be found in EVM or GMM. Thus, this method should not be difficult
for all those who know and use AHP. HRE, unlike AHP, does not force
the user to abandon the actual measured values associated with given
alternatives in favor of the values calculated as a result of the
ranking procedure. Once measured or calculated, the selected quantities
become reference values in this model and are not subject to further
change. If this behavior is desired in a given application, the HRE
method will work here. 

We believe that HRE fills a specific gap and can prove its usefulness
wherever tangible and intangible criteria need to be combined within
a single hierarchical model. Therefore, it might interest a range
of practitioners who may feel a little awkward when comparing well-measured
values such as price, length, or weight, and theoreticians studying
the properties of new decision-making methods.

\section{Literature}

\bibliographystyle{plain}
\bibliography{papers_biblio_reviewed}

\end{document}